\documentclass[preprint,authoryear,12pt]{elsarticle}

\usepackage{graphicx} 

\usepackage{amssymb}
\usepackage{amsmath}
\usepackage{stmaryrd}

\usepackage{algorithm}
\usepackage{algorithmic}
\usepackage{color}
\usepackage[bookmarks=true, bookmarksnumbered=true, bookmarksopen=true,
  unicode=true, colorlinks=true,
  linkcolor=blue,citecolor=blue,filecolor=blue,urlcolor=blue,
  pdfstartview=FitH
]{hyperref}
\usepackage{url}
\usepackage{xcolor}

\DeclareMathOperator*{\argmax}{arg\,max}

\journal{Computer Speech and Language}

\begin{document}

\begin{frontmatter}



\title{Multiple topic identification in\\human/human conversations}


\author[lia]{X.~Bost}
\ead{xavier.bost@univ-avignon.fr}

\author[lia]{G.~Senay\corref{cor}}
\ead{gregory.senay@gmail.com}

\author[lia]{M.~El-B{\`e}ze}
\ead{marc.elbeze@univ-avignon.fr}

\author[lia,mcgill]{R.~De Mori}
\ead{rdemori@cs.mcgill.ca}

\address[lia]{LIA, University of Avignon, 339 chemin des Meinajari{\`e}s, 84000 Avignon, France}
\address[mcgill]{McGill University, School of Computer Science, Montr{\'e}al, Qu{\'e}bec, Canada}

\cortext[cor]{Corresponding author}

\begin{abstract}

\noindent \textcolor{red}{\textbf{Cite as:}\\X.~Bost, G.~Senay, M.~El-B\`eze, R. De Mori.\\\href{https://www.sciencedirect.com/science/article/pii/S0885230815000352?via\%3Dihub}{Multiple topic identification in human/human conversations.}\\Computer Speech \& Language, 2015, 34 (1), pp.18-42.\\doi: \href{https://doi.org/10.1016/j.csl.2015.03.006}{10.1016/j.csl.2015.03.006}}

\vspace{2mm}

The paper deals with the automatic analysis of real-life telephone conversations between an agent and a customer of a customer care service (\textsc{ccs}). The application domain is the public transportation system in Paris and the purpose is to collect statistics about customer problems in order to monitor the service and decide priorities on the intervention for improving user satisfaction.

Of primary importance for the analysis is the detection of themes that are the object of customer problems. Themes are defined in the application requirements and are part of the application ontology that is implicit in the \textsc{ccs} documentation.

Due to variety of customer population, the structure of conversations with an agent is unpredictable. A conversation may be about one or more themes. Theme mentions can be interleaved with mentions of facts that are irrelevant for the application purpose. Furthermore, in certain conversations theme mentions are localized in specific conversation segments while in other conversations mentions cannot be localized. As a consequence, approaches to feature extraction with and without mention localization are considered.  

Application domain relevant themes identified by an automatic procedure are expressed by specific sentences whose words are hypothesized by an automatic speech recognition (\textsc{asr}) system. The \textsc{asr} system is error prone. The word error rates can be very high for many reasons. Among them it is worth mentioning unpredictable background noise, speaker accent, and various types of speech disfluencies.

As the application task requires the composition of proportions of theme mentions, a sequential decision strategy is introduced in this paper for performing a survey of the large amount of conversations made available in a given time period. The strategy has to sample the conversations to form a survey containing enough data analyzed with high accuracy so that proportions can be estimated with sufficient accuracy.

Due to the unpredictable type of theme mentions, it is appropriate to consider methods for theme hypothesization based on global as well as local feature extraction. Two systems based on each type of feature extraction will be considered by the strategy. One of the four methods is novel. It is based on a new definition of density of theme mentions and on the localization of high density zones whose boundaries do not need to be precisely detected.

The sequential decision strategy starts by grouping theme hypotheses into sets of different expected accuracy and coverage levels. For those sets for which accuracy can be improved with a consequent increase of coverage a new system with new features is introduced. Its execution is triggered only when specific preconditions are met on the hypotheses generated by the basic four systems.

Experimental results are provided on a corpus collected in the call center of the Paris transportation system known as \textsc{ratp}. The results show that surveys with high accuracy and coverage can be composed with the proposed strategy and systems. This makes it possible to apply a previously published proportion estimation approach that takes into account hypothesization errors.
\end{abstract}


\end{frontmatter}


\section{Introduction}

In recent years, there has been an increasing research interest in the analysis of human/human spoken conversations. Advances in this research area are described in \citep{tur2011spoken} and \citep{tur2011human}. A scientifically interesting and practically important component of this research is topic identification for which an ample review of the state of the art can be found in \citep{hazen2011topic}. In spite of the relevant progress achieved so far, it is difficult to reliably identify multiple topics in real-life telephone conversations between casual speakers in unpredictable acoustic environments. Of particular interest are call-center conversations in which customers discuss problems in specific domains with an advisor. This is the case of the application considered in this paper. The purpose of the application is to monitor the effectiveness of a customer care service (\textsc{ccs}) of the \textsc{ratp} Paris transportation system by analyzing real-world human/human telephone conversations in which an agent solicits a customer to formulate a problem and attempts to solve it. The  application task is to perform surveys of customer problems in different time periods. Proportions of problem themes computed with the survey data are used for monitoring user satisfaction and establishing priorities of problem solving interventions. Of primary importance for this task is the ability to automatically select, in a given period of time, a sufficiently large sample of conversations automatically analyzed with high accuracy. Application relevant information for evaluating proportions of problem items and solutions is described in the application requirements making evident, not formally defined, but useful, speech analytics. Themes mentioned in the application documentation appear to be the most important and general classes representing concerns that the survey has to address. Themes appear to be an adequate semantic representation of the types of problems in the application ontology that is inferred from the documentation.

Agents follow a pre-defined protocol to propose solutions to user problems about the transportation system and its services. In order to evaluate proportions of problem types, agents compile conversation reports following a domain ontology. Due to time constraints, reports compiled by the agents cover a small proportion of conversations. Furthermore, compiled reports are often incomplete and error-prone. Thus, an automatic classification of conversation themes would be useful for producing a quantity of accurate reports sufficient for performing a reliable survey. 

The achievement of acceptable automatic solutions has to overcome problems and limits of known spoken language processing and interpretation systems. A fully automatic system must include an automatic speech recognition (\textsc{asr}) module for obtaining automatic transcriptions of the conversations. The acoustic environment on these conversations is unpredictable with a large variety of noise types and intensity. Customers may not be native French speakers and conversations may exhibit frequent disfluencies. The agent may call another service for gathering information. This may cause the introduction of different types of non-speech sounds that have to be identified and discarded. For all these reasons, the word error rate (\textsc{wer}) of the \textsc{asr} system is highly variable and can be very high. Conversations to be analyzed may be about one or more domain themes. Mentions of relevant themes may be interleaved with mentions of irrelevant comments. Mentions of multiple themes may appear in well separable conversation segments or they may be diffused in zones of a conversation that are difficult to localize. Moreover, mentions can be incomplete or highly imprecise involving repetitions, ambiguities, linguistic and pronunciation errors. Some mentions of an application relevant theme may become irrelevant in certain contexts. For example, a customer may inquiry about an object lost on a transportation mean that was late. In such a case, the loss should be considered as a more relevant theme than the traffic state delay.

In order to properly approach the problems in the above mentioned variety of scenarios and difficulties, methods relying on evaluating the evidence of each feature in the whole conversation should be combined with methods that evaluate feature evidence on specific, automatically detected, conversation segments. In fact, depending on the customer speaking style, mentions of the same pair of themes can be diffused in large portions of a conversation with not clearly defined location for each theme, or they may appear in other conversations in well-localized segments. Thus, it appears useful to consider algorithmic approaches for diffuse theme mentions and other approaches for mentions in specific, localized segments. Furthermore, in order to reduce the effect of \textsc{asr} and classification errors, it is proposed in this paper to investigate the possibility of using at least two different approaches for each of the two possible types of mentions. One of these methods is novel and is introduced, together with a review of the others, in section~\ref{sec:archi}, while the corpus used for the experiments is described in section~\ref{sec:appli}.
 
In order to obtain reliable surveys with error-prone approaches, conditions have to be automatically identified for selecting reliably annotated samples and for possibly triggering suitable refinements.  An effective sequential theme hypothesization strategy will be introduced in section~\ref{sec:strat}. A component of the strategy performs refinements consisting in recovering deletion or insertion errors in multiple theme hypotheses having a correct theme already hypothesized. The strategy also includes a new theme evaluation process to compensate some specific classification errors that may be caused by \textsc{asr} errors. The execution of this process is restricted to specific situations. The process action performs semantically coherent, compositions or decompositions of already generated theme hypotheses. New features are used in this process. They are designed to characterize diffuse theme mentions and are extracted from word lattices as described in \citep{wintrode2009techniques}, \citep{hazen2011mce}. A moderate amount of human compiled knowledge is conceived to establish preconditions inspired by the application ontology for the application of the refinement process. The components of the theme hypothesization strategy are conceived to have low linear time complexity making their execution faster that the \textsc{asr} process. The experimental results reported in sections~\ref{sec:ind_res} and \ref{sec:strat} show that topic identification errors increase with the disagreement among the four systems. Nevertheless, reliable surveys can be obtained with high annotation accuracy in a high proportion of the available conversations with a sequential decision strategy that progressively applies specific decision criteria and features from \textsc{asr} generated lattices of word hypotheses.

\section{Related work}

As mentioned in the topic identification review presented in \citep{hazen2011topic} topic identification is performed with supervised and unsupervised classification methods. It is worth mentioning among them, especially for text documents, decision trees \citep{cohen1999context}, naive Bayes classifiers \citep{li1998classification}, other probabilistic classifiers \citep{lewis1994sequential}, support vector machines \citep{joachims1998text}, k-nearest neighbors and example-based classifiers \citep{yang1997comparative}.

Popular features used by the classifiers are terms, for each of which measures such as term frequency (\textsc{tf}), inverse document frequency (\textsc{idf}), and \textsc{tf}-\textsc{idf} combinations are computed. A review on the unsupervised selection of terms such as keywords and word chunks can be found in \citep{chen2010automatic}. These features can be enriched with variations of left and right context and with parts of speech (\textsc{pos}) for each selected word. Other feature selection methods are discussed in \citep{yang1997comparative}.

In order to reduce the dimensionality of the feature space, latent semantic analysis (\textsc{lsa}) \citep{bellegarda2000exploiting}, its probabilistic version (p\textsc{lsa}) and latent Dirichlet allocation (\textsc{lda}) have been used for topic identification. In \citep{chien2008adaptive} a naive Bayes classifier based on adaptive \textsc{lsa} is proposed for topic identification in spoken documents. With \textsc{lda} \citep{blei2003latent}, \citep{wintrode2011using}, the features representing a document are probabilities of the presence of hidden topics in a document. Hidden topics are determined by a specific learning process and differ from the topics to be identified. Approaches based on the assumption that a document is represented by a feature vector that is a combination of a small number of topic vectors are reviewed and compared in \citep{min2013joint}.

Recently \citep{morchid2014vector}, it has been observed that theme identification accuracy may have large variations for different choices and dimensions of hidden topic spaces. Stable and superior performance has been achieved by introducing a new method called \textit{c}-vector proposed for integrating features obtained in a large variety of these spaces.

Topic segmentation in spoken conversations is reviewed in \citep{purver2011topic}. An evaluation of coarse-grain discourse segmentation can be found in \citep{niekrasz2010unbiased}. Interesting solutions have been proposed for linear models of nonhierarchical segmentation. Some approaches propose inference methods for selecting segmentation points at the local maxima of cohesion functions. Some functions use features extracted in each conversation sentence or in a window including a few sentences. Some search methods detect cohesion in a conversation using language models and some others consider hidden topics shared across documents. These aspects are discussed for text processing in \citep{eisenstein2008bayesian} together with a critical review on lexical cohesion. In \citep{balchandran2010techniques} techniques for topic detection have been proposed for obtaining topic specific language models (\textsc{lm}s) used for re-scoring \textsc{asr} results. 

Multi-label text classification is discussed in \citep{de2009tutorial} and \citep{tsoumakas2007multi} for large collections of text documents. In \citep{de2009tutorial} a technique, called \textit{creation}, is proposed. It consists in creating new composite labels for each association of multiple labels assigned to an instance. 
An important contribution with useful discussions and a thorough review on learning semantic structures from in-domain text document can be found in \citep{chen2010learning}. A generative model of content structure is proposed based on an explicit representation of discourse constraints on regularities that are important for topic selection.

This paper adapts concepts found in the recent literature to the design of an automatic process for reliably annotating and selecting conversations for a \textsc{ccs} survey. A new sequential decision strategy is proposed for this purpose. It uses results of four different classification systems conceived for searching mentions of theme relevant features in an entire conversation or in specific segments of it. The motivation for considering these two possibilities is the unpredictable language style of casual real-world application users. One of the four systems uses a new method for extracting features from discourse segments. Rather than building segments by collapsing sentences or paragraphs or by identifying suitable segment bounds, the proposed approach looks for zones of a conversation dense of theme relevant features and uses these data for classification. This is motivated by the observation that often customers tend to introduce their problem with factual descriptions that are irrelevant for the task and agents tend to formulate a solution with a verbose introduction followed by a concise formal explanation dense of theme specific words and phrases. 

In \citep{wintrode2009techniques}, a method is introduced for performing a supervised classification of a spoken document into a single topic class. Features for the classification are posterior probabilities of word hypotheses computed over an entire conversation and obtained from word lattices generated  by an \textsc{asr} system. These lattice features are also applied in \citep{hazen2011mce} to supervised topic classification with minimum error classifiers. Hidden topic features obtained with \textsc{lda} have been combined with other features in support vector machine (\textsc{svm}) classifiers to assign spoken documents to relevance classes \citep{wintrode2011using} using \textsc{asr} word lattice hypotheses.

The strategy proposed in this paper is based on an application independent criterion for composing conversations sets of increasing size with limited decrease in automatic annotation accuracy. A specific strategy process attempts to alleviate the effects of possible \textsc{asr} errors by using specific word and distance bigram features extracted from the lattice of word hypotheses generated by an \textsc{asr} system. These bigrams are selected to express distance relations between facts and actions that characterize each theme.
 
\section{The application domain and the corpus used for the experiments}

\label{sec:appli}

The application task considered in this paper is the automatic annotation of conversations between a customer and the agent of a \textsc{ccs}. Application requirements establish that the conversations have to be annotated in terms of application domain semantic contents. Essential contents are themes belonging to the following set (between parentheses the abbreviations used by the agents): \\

$\mathbb{T} := \ $\{\textit{itinerary \textsc{(itnr)}, lost and found \textsc{(objt)}, time schedules \textsc{(horr)}, transportation card \emph{navigo} \textsc{(nvgo)}, traffic state \textsc{(etfc)}, fares \textsc{(tarf)}, contraventions \textsc{(pv)}, transportation card \emph{vgc} \textsc{(vgc)}, out of domain \textsc{(trsh)}}\} \\

Identifying these themes is the purpose of the research described in this paper. Conversations may be about more than one theme and the co-presence of related themes appears to be essential to characterize the customer problem and to adequately monitor the time evolution of problem proportions in order to assess the increase or decrease of mentions of user concerns. Such an assessment can be made with surveys containing conversation samples. A method for obtaining acceptable estimations of proportion variability in time using surveys with samples that may contain annotation errors can be found in \citep{camelin2009error}. Starting with themes annotated with sufficient accuracy, more specific semantic information can be extracted and used to provide more details in conversation reports. These details can be used for other tasks such as the evaluation of the impact of strikes at specific dates. This aspect will be considered in future work.

Conversations are made available in successive time periods. Samples taken from different periods of a year have been selected to form a corpus of 1,654 telephone conversations collected at the call center of the public transportation service in Paris (\textsc{ratp}). The corpus has been manually transcribed after removing any relation to customer identity and other confidential information. In order to perform experiments with a fully automatic system, the corpus was split into three corpora, namely a train, a development and a test set containing respectively 880, 196 and 576 conversations. The sizes of the corpora in terms of conversations annotated with one or more theme labels are shown in Table~\ref{Decoda_Data}. The table includes, for each set, the number of conversations annotated with a single theme label (\#mono-label), the number of conversations annotated with multiple theme labels (\#multi-label), the label cardinality (\textit{i.e.} the average number of labels by conversation, denoted here \textit{label card.}) and the percentage of conversations annotated with multiple labels.

\begin{table}[h]
  \caption{\label{Decoda_Data}{\it Decoda corpus - Statistical data}}
  \vspace{2mm} \centering
  \begin{tabular}{|c||c|c|c||c|}
    \hline
    Corpus 		& Train	& Dev 	& Test	& Total	\\ \hline
    \#			& 880	& 196 	& 576	& 1,654	\\ \hline
    \#mono-label	& 744	& 146 	& 424	& 1,314	\\ \hline
    \#multi-label 	& 136	& 50    & 154 	& 340	\\ \hline
    label card. 	& 1.16	& 1.29 	& 1.29	& 1.22	\\ \hline
    \%multi 		& 15.45	& 25.51	& 26.64	& 20.56	\\ \hline
  \end{tabular}
\end{table}

Conversations have been independently annotated by three human annotators in terms of the main theme and possibly additional themes. When the annotators disagree, a consensus was found after discussion. The corpus with manual transcriptions and annotations will be made publicly available at the end of the \textsc{decoda} project pending the authorization of the data owners. In order to develop an effective strategy, four initial different systems have been considered and developed for automatic annotation. All the four systems use the same type of features extracted from the 1-best sequence of word hypotheses generated by an \textsc{asr} component described in \citep{linares2007lia}. The \textsc{asr} system is based on triphone acoustic hidden Markov models (\textsc{hmm}) with mixtures of 230,000 Gaussian distributions. Model parameters were estimated with maximum \textit{a posteriori} probability (\textsc{map}) adaptation of 150 hours of speech in telephone bandwidth with the data of the train set. A 4-gram language model (\textsc{lm}) was obtained by adapting with the transcriptions of the train set a basic \textsc{lm} whose parameters have been estimated with other corpora. An initial set of experiments was performed with this system. The results show a \textsc{wer} on the test set of 57\% (52\% for agents and 62\% for customers). These high error rates are mainly due to speech disfluencies and to adverse acoustic environments for some conversations when, for example, customers call from train stations or noisy streets with mobile phones. Furthermore, the signal of some sentences is saturated or of low intensity due to the distance between speakers and phones. Such difficult acoustic conditions have a significant impact on the density of the lattices with peaks of 2,600 links per second.

Two systems called \textsc{density} and \textsc{hmm} segment a conversation and attempt to hypothesize theme mentions in segments. The other two systems called \textsc{cosine} and \textsc{poisson} generate annotation hypotheses by cumulating scores for each feature in the entire conversation. 

The \textsc{asr} component generates a lattice of word hypotheses using a vocabulary of a fixed number of words obtained from the content of the database of the application domain and words observed in the train corpus. The test data may contain out-of-vocabulary (\textsc{oov}) words not observed in the train set. It was observed that the train corpus vocabulary contains 7,920 words while the test corpus contains 3,806 words, only 70.8\% of them occur in the train corpus and the model perplexity is 82 for the \textsc{dev} set and 86 for the \textsc{test} set. In spite of possibly frequent \textsc{asr} errors, it is worth investigating the possibility of performing reliable theme surveys considering that conversations contain mentions to rich sets of theme relevant facts, entities and actions and many small sets of them are sufficient conditions for theme identification.

Theme identification may use the most likely sequence (1-best sequence) of word hypotheses as features for theme identification. Using features from the lattice of word hypotheses may only reduce the effect of some errors in the 1-best sequence of words in the vocabulary, but may also introduce confusion in the classification decision performed by the theme identification strategy.

In addition to the \textsc{oov} problem, theme identification may also confront the code-switching problem occurring when speakers alternate between language varieties. An analysis of the train set shows that the code-switching problem is not frequent and appears only in some customer segments. This is explained by the fact that the agent has to follow a well-defined problem solving protocol and tends to drive the conversation according to it. Furthermore, most of the varieties and use of \textsc{oov} words relate to descriptions of facts that are irrelevant for the application purpose. In any case, in order to model possible code-switching effects, two systems, namely \textsc{density} and \textsc{hmm}, have been conceived to analyze conversation segments primarily for using localized features to model possible variations in theme mentions.

For the sake of comparison, the four systems perform classifications with different methods using the same features extracted from the 1-best sequence of word hypotheses.

As the four systems are based on different algorithms, it is expected that they make different types of classification errors and tend to produce different annotations mostly when the features are affected by drastic \textsc{asr} errors. 

Parameter estimation for each system is based on the minimization of classification errors and comparisons of scores with thresholds for making decisions. For this reason, the consensus among the four systems is expected to be a good confidence index of the classification results. Consensus is thus used for an initial selection of the conversations to be inserted in the survey.

In order to find, in specific conditions of weak consensus, additional conversations to be placed in the survey, the strategy is refined by a conditional execution of a fifth new process based on a new method and new features. These features are specific for each theme and are extracted from the lattice of word hypotheses generated by the \textsc{asr} component. Details of these features will be provided later on with the description of the refinement strategy.

\section{Multiple theme hypothesization architecture}

\label{sec:archi}

An architecture for multiple theme hypothesization has been conceived for integrating the results of different systems implementing different global and segmental approaches for extracting mentions of application relevant themes.

\begin{figure}[h]
  \centering
  \includegraphics[width=90mm]{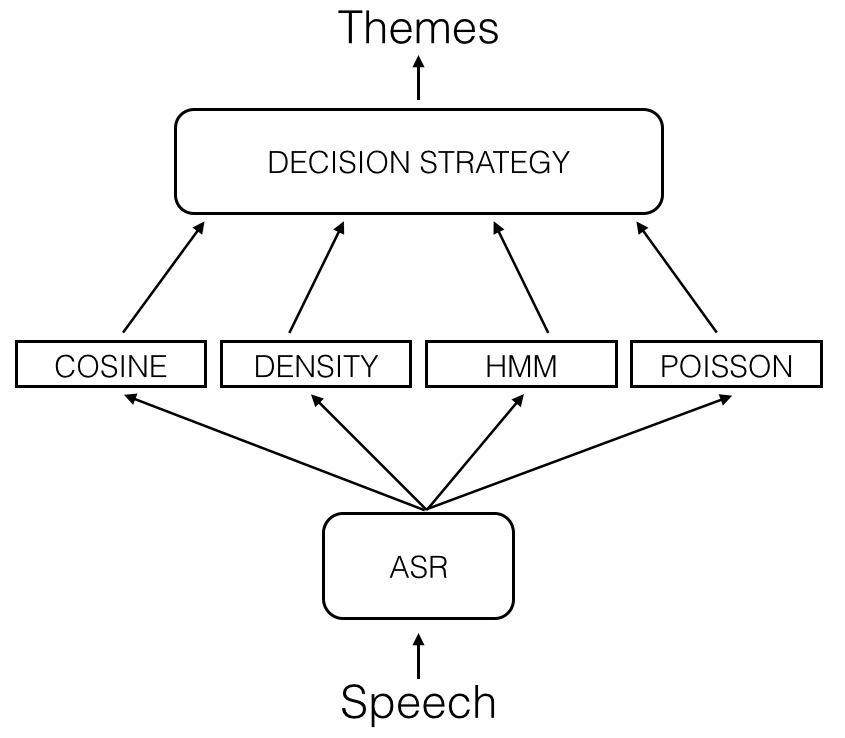}
  \caption{\label{System architecture} 
    {\it Multiple theme hypothesization architecture}}
\end{figure}

The architecture has the scheme shown in Figure~\ref{System architecture} and is based on three components, namely an \textsc{asr} system that generates word hypotheses, a theme hypothesization component that contains four systems whose results are further processed by a decision making strategy.

The four systems use the most likely (1-best) sequence of word hypotheses generated by the same \textsc{asr} component. Different features extracted from the lattice of word hypotheses are available. They are processed by a fifth system whose output can be used by the strategy for making decisions when there are doubts about the confidence of the outputs of the four other systems. The four basic systems use the same set of features made of words selected based on their purity, word classes and short distance word bigrams.

Theme hypotheses are generated by systems called \textsc{cosine} and \textsc{poisson} using theme mentions detected anywhere in a conversation, while \textsc{hmm} and \textsc{density} localize theme mentions in specific positions of a conversation and make decisions based on features extracted around these positions. The four systems are introduced below.

A short description and preliminary results obtained with \textsc{cosine} and \textsc{density} have been presented in \citep{bost2013multiple}. As \textsc{density} is a novel method, it will be described here in more details. 

The features used by the four systems contain a selection of 7,217 words in the \textsc{asr} lexicon, a few word classes such as times and prices, and word bigrams. Distance bigrams made of pairs of words distant a maximum of two words are also used. A reduced feature set $V_f$ is obtained by selecting features based on their purity and coverage. Purity of a feature $f$ is defined with the Gini criterion as follows:

\begin{equation}
    G(f) = \sum_{t \in \mathbb{T}} \mathbb{P}^2(t|f)
    = \sum_{t \in \mathbb{T}} \left ( \frac{df_{t}(f)}{df_{\mathbb{C}}(f)} \right ) ^2
\end{equation}

where $df_{\mathbb{C}}(f)$ is the number of conversations of the train set $\mathbb{C}$ containing feature $f$ and $df_{t}(f)$ is the number of conversations of the train set containing feature $f$ and annotated with theme $t$.

Considering a conversation as a document, a score $w_t(f)$ for feature $f$ is defined, using all the samples of $ t$ in the train set, as follows:

\begin{equation}
  w_t(f) = df_{t}(f) . idf^2(f) . G^2(f)
\end{equation}

where $idf(f)$ is the inverse document frequency for feature~$f$.

\subsection{Cosine}

\label{subsec:cosine}

The \textsc{cosine} system generates a theme hypothesis with a classification process based on a global cosine measure of similarity  $\text{sc}(y, t)$ between a conversation $y$ and a theme $t$.
Representing $y$ by a vector $\mathbf{v_y}$ and $t$ by a vector $\mathbf{v_t}$, this measure of similarity is computed as follows:

\begin{equation}
  \text{sc}(y, t) = \cos(\widehat{\mathbf{v_y}, \mathbf{v_t}})
    = \frac{\sum_{f \in y \cap t}w_y(f).w_t(f)}{\sqrt{\sum_f w_y(f)^2 . \sum_f w_t(f)^2}}
\end{equation}

where $w_y(f)$ is a score for feature $f$ in conversation $y$.

Let $\gamma_1(y)$ be the set of themes discussed in $y$. The following decision rule for automatically annotating conversation $y$ with a theme class label $t$ is initially applied:

\begin{equation}
  t \in \gamma_1(y) \Longrightarrow \text{sc}(y, t) \geqslant \rho.
  \text{sc}(y, \hat{t})
\end{equation}

where $\hat{t}:= \argmax_{t' \in \mathbb{T}}\text{sc}(y, t')$; and $\rho \in [0; 1]$ is an empirical parameter whose value is estimated by experiments on the development set. If the score of $\hat{t}$ is too low, then the application of the above rule is not reliable. To overcome this problem, the following additional rule is introduced:

\begin{equation}
  t \in \gamma_1(y) \Longrightarrow \text{sc}(y, t) \geqslant v 
  \sum_{t' \in \mathbb{T}} \text{sc}(y, t')
\end{equation}

where $v \in [0; 1]$ is another parameter whose value is estimated by experiments on the development set.

The optimal value for $\rho$, the proportion of the highest score required for assigning themes to a conversation, has been evaluated to $\hat{\rho} = 0.69$, and the optimal value threshold $v$ to $\hat{v} = 0.16$.
 
\subsection{Density}

\label{subsec:density}

Rather than representing a conversation with a \textit{bag of words} whose order is irrelevant, the \textsc{density} system analyzes sequences of transcribed words for extracting features at specific positions in a sequence. A conversation containing $N$ words is seen, including spaces, as a finite sequence $(p_1, ..., p_n$) of $n$ positions (with $n = 2N -1$). Features are described taking into account the position where they are extracted. The $k$-th word unigram in the feature sequence representation is located at position $p_{2k-1}$, the $k$-th bigram is at  position $p_{2k}$, the $k$-th distant bigram with one intermediate unigram is at position $p_{2k+1}$.

The contribution to theme $t$ of the features at the $i$-th position in a conversation is:

\begin{equation}
  w_t(p_i) = \frac{1}{\parallel \mathbf{v_t} \parallel} \sum_{f \in \tau_{p_i}} w_t(f) \ \  (i = 1, ..., n)
\end{equation}

where $\tau_{p_i}$ is the set made of the features (unigram and associated bigrams) located at position $p_i$ in a conversation.

A thematic density $d_{t}(p_i)$ of theme $t$ is associated with position $p_i$ and is defined as follows:

\begin{equation}
  d_{t}(p_i) = \frac{\sum_{j=1}^n
    \frac{w_t(p_j)}{\lambda^{d_j}}}{\sum_{j=1}^n
    \frac{1}{\lambda^{d_j}}} \ (i = 1, ..., n)
  \label{eq:ED1}
\end{equation}

where $\lambda \geqslant 1$ is a parameter of sensitivity to proximity whose value is estimated by experiments on the development set and $d_j:= |i - j|$ is the distance in a conversation between positions $p_i$ and $p_j$.

\subsubsection{Theme density computation}

According to equation~\eqref{eq:ED1}, the density for theme $t$  at the $i$-th position of a conversation is obtained by adding the contributions of the features located at that point and the contributions of each surrounding $(i \pm d)$ position weighted according to the distance $d$. Such a computation requires two nested loops over the $n$ conversation positions, resulting in a quadratic time complexity.

In order to reduce computation costs, recurrence relations have been introduced in the expression of equation~\ref{eq:ED1}, resulting in a linear computation algorithm.

This algorithm, as well as the recurrence formulas it relies on, are fully specified in~\ref{sec:dens_comp}.

\subsubsection{Conversation skeleton}

The thematic density at a specific conversation position is used to construct a thematic skeleton by plotting the density of theme hypotheses as function of a position measured in number of words (including spaces) preceding the position.

An example of thematic skeleton for a particular conversation is reported in~\ref{sec:conv_skel}. A similar example can be found in~\citep{bost2013multiple}.

\subsubsection{Decision making}

A theme $t$ is considered as discussed in a conversation $y$ if it has dominant density at a position of the conversation (rule \eqref{eq:RL1}) and if the sum of its densities in positions where it is dominant exceeds an empirically determined threshold (rule \eqref{eq:RL2}):

\begin{equation}
  t \in \gamma_2(y) \Longrightarrow \exists i \in \llbracket {1};{n}
  \rrbracket: \forall t' \in \mathbb{T}, \ d_{t}(p_i) \geqslant
  d_{t'}(p_i)
  \label{eq:RL1}
\end{equation}

\begin{equation}
  t \in \gamma_2(y) \Longrightarrow \sum_{i \in I} d_{t}(p_i) \geqslant
  v \sum_{j=1}^n d_{t_j}(p_j) 
  \label{eq:RL2}
\end{equation}

where $\gamma_2(y)$ is the set of themes hypothesized for conversation $y$; $v \in [0, 1]$ is a parameter whose value is estimated by experiments on the development set; $t_j$ is the theme of dominant density at the $j$-th position in the conversation; and $I:= \{i \in \llbracket {1};{n} \rrbracket \ | \ \forall t' \in \mathbb{T}, \ d_t(p_i) \geqslant d_{t'}(p_i) \}$.

\subsection{\textsc{hmm}}

\textsc{hmm}s have been widely used for topic identification (see, for example, \citep{mittendorf1994document} and \citep{wallace2013generative} for a recent application). 

For each theme $t_k$, a language model $\textsc{lm}_k$ is obtained with the train set. The language model $\textsc{lm}_k$ is made of unigram probabilities and of probabilities $\mathbb{P}_k(w h)$, where histories are obtained from chunks automatically selected with the same procedure used in \citep{maza2011use}. Conditional probabilities $\mathbb{P}(t_i|t_j)$ are estimated from the conversations of the train set annotated with two themes. These probabilities are used together with probabilities $\mathbb{P}(t_k)$ that there is a unique theme in a conversation. 

A network is then constructed by connecting in parallel theme \textsc{hmm}s modeling single theme conversations and sequences of these \textsc{hmm}s for modeling multiple theme conversations. Probabilities of having a unique theme or pairs of themes are associated with each branch of the network. Word generation probabilities of these models are provided by the theme \textsc{lm}s. The Viterbi algorithm is then applied using the network as model and the words of the 1-best sequence of conversation $y$ as observations for computing the highest probability among the following candidates $\mathbb{P}(y|t_k)$, $\mathbb{P}(y |t_mt_j)$, for every unique theme $(\forall k)$ and for every pair $(\forall  (m, j))$. When the hypothesis of multiple themes is predominant, segments expressing pairs of hypothesized themes are hypothesized by the search algorithm, together with a score for each segment. In this way, segments of theme mentions are hypothesized using models of short term sequential constraints.


\subsection{\textsc{poisson}}

The optimal hypothesis $\hat{t} $ consisting of a single theme or a sequence of themes is obtained with the following equation:

\begin{equation}
    \hat{t} = \argmax_t (\mathbb{P}(t|y)) = \argmax_t (\log(\mathbb{P}(y|t)) + \log(\mathbb{P}(t)))
  \label{eq:EQ1}
\end{equation}

Given the relatively small size of the available corpus, the Poisson law is well suited to take into account the sparse distribution of features $f$ in a theme $t$. The approach followed for finding the results reported in this paper is inspired by \citep{bahl1988obtaining}. Relying on the histogram estimate $F$ of feature occurrences, with mean $\mu_{f,t}$, a topic model has been conceived based on the following probability:

\begin{equation}
  \mathbb{P}(F_f = r|t) = \frac{\text{e}^{-\mu_{f,t}} \mu_{f,t}^r}{r!}
\end{equation}

According to \citep{bahl1988obtaining} it is assumed that the feature frequencies have independent Poisson distribution and $\log(\mathbb{P}(y|t))$  can be estimated as follows:

\begin{equation}
  \log(\mathbb{P}(y|t)) \simeq \sum_f (F_f \log(\mu_{f,t}) - \mu_{f,t} - \log(F_f!))
\end{equation}

The last logarithm in the summation is independent of $t$ and can be ignored leading to:

\begin{equation}
  \hat{t} = \argmax_t (\log(\mathbb{P}(t)) + \sum_f (F_f \log(\mu_{f,t}) - \mu_{f,t}))
\end{equation}

\section{Individual evaluation of the four systems for theme hypothesization}

\label{sec:ind_res}

\subsection{Performance measures}

\label{subsubsec:metrics}

The proposed approaches have been evaluated following procedures discussed in \citep{tsoumakas2007multi} with measures used in Information Retrieval (\textsc{ir}) and accuracy as defined in the following for a corpus $\mathbb{X}$ and a decision strategy $\gamma$.

\medskip

\textbf{Recall}:

\begin{equation}
  R(\gamma, \mathbb{X}) = \frac{1}{|\mathbb{X}|} \sum_{y \in\mathbb{X}}
  \frac{|\gamma(y) \cap M(y)|}{|M(y)|}
\end{equation}

where $M(y)$ indicates the set of themes manually annotated for conversation $y$ and $\gamma(y)$ the set of themes automatically hypothesized for the same conversation.

\medskip

\textbf{Precision}:

\begin{equation}
  P(\gamma, \mathbb{X}) = \frac{1}{|\mathbb{X}|} \sum_{y \in\mathbb{X}}
  \frac{|\gamma(y) \cap M(y)|}{|\gamma(y)|}
\end{equation}

\medskip

\textbf{F-score}:

\begin{equation}
  F(\gamma, \mathbb{X}) = \frac{2P(\gamma, \mathbb{X})R(\gamma,
    \mathbb{X})}{P(\gamma, \mathbb{X})+R(\gamma, \mathbb{X})}
\end{equation}

\medskip

\textbf{Accuracy}:\\

The traditional way of computing the accuracy (denoted hereafter $acc_1$) is defined as:

\begin{equation}
  acc_1(\gamma, \mathbb{X}) = \frac{1}{|\mathbb{X}|} \sum_{y \in\mathbb{X}}
  g(\gamma(y), M(y))
  \label{eq:ACC1}
\end{equation}

where:

\begin{equation}
  g(\gamma(y), M(y)) = 
  \begin{cases}
    1 & \text{ if } \gamma(y) = M(y) \\
    0 & \text{ otherwise}
  \end{cases}
  \label{eq:ACCAUX}
\end{equation}

For multiple topic identification, it is also interesting to compute the accuracy in another way (denoted $acc_2$), as proposed by \citep{tsoumakas2007multi}:

\begin{equation}
  acc_2(\gamma, \mathbb{X}) = \frac{1}{|\mathbb{X}|} \sum_{y \in\mathbb{X}}
  \frac{|\gamma(y) \cap M(y)|}{|\gamma(y) \cup M(y)|}
\end{equation}

In the following if only \textit{accuracy} is mentioned, it has to be assumed to refer to $acc_1$.

\subsection{Results}

\label{subsec:res}

The four systems \textsc{density}, \textsc{hmm}, \textsc{poisson}, and \textsc{cosine} have been separately evaluated on the development and test sets. A linear interpolation of the results obtained with the four systems has also been evaluated and results are reported in the line labelled \textit{comb}. As the \textsc{dev} set is too small for estimating the weights of the linear combination, equal weights have been used for the scores of the four systems.

As the parameters of the four systems are estimated for separately optimizing the accuracy of each system, a linear combination of decision scores of each system is not as efficient as consensus. The reason is that consensus is based only on each system decision that is reasonable to consider as the most reliable indicator of the evidence ascribed by each system to the theme hypotheses it generates after classifying with a specific approach an entire conversation.

The results are reported for the \textsc{dev} set in Tables~\ref{table_res_dev_man} and~\ref{table_res_dev_asr}, and for the \textsc{test} set in Tables~\ref{table_res_test_man} and~\ref{table_res_test_asr}. Label \textsc{man} refers to automatic theme annotation of manual transcriptions, while label \textsc{asr} refers to automatic theme annotation of \textsc{asr} system transcription.

For the sake of comparison, the results obtained with the proposed classification approaches have been compared with those obtained with a support vector machine (\textsc{svm}) using the same features (unigrams and bigrams with possible gap of one word) and a linear kernel. Denoting by $\mathbb{X}$ the set of conversations, a binary classifier $\gamma_k : \mathbb{X} \rightarrow \{t_k, \overline{t_k} \}$ is defined for every theme $t_k \in \mathbb{T}$. For every pair $(y, t_k)$ containing the conversation $y$ and the $k$-th theme, a score is computed by using this $k$-th classifier. The candidate theme hypotheses for a conversation are those whose score is in an interval corresponding to an empirically determined proportion of the highest one. In addition to that, the hypothesis with the highest score must be above a threshold empirically determined for this purpose.

The toolkit used to perform \textsc{svm}-based categorization, \textsc{svm}\textsuperscript{light}, is described in \citep{joachims2002learning}, \citep{joachims1999making} and \citep{joachims2002optimizing}.

\begin{table}[h]
  \caption{\label{table_res_dev_man} {\it Results obtained for the manual transcriptions of the \textsc{dev} set with the \textsc{svm}, the four systems and their linear combination (confidence interval = 0.03)}}
  \vspace{2mm} \centering
  \begin{tabular}{|c|c|c|c|c|c|}
    \hline
    \textsc{dev} & \multicolumn{5}{|c|}{\textsc{man}} \\
    \cline{2-6}
    & $acc_1$ & $acc_2$ & \textbf{prec.} & \textbf{rec.} & \textbf{F-sc.} \\
    \hline
    \hline
    \textbf{svm.} & 0.62 & 0.78 & 0.87 & 0.87 & 0.87 \\
    \hline
    \hline
    \textbf{cos.} & 0.72 & 0.85 & 0.95 & 0.88 & 0.92 \\
    \hline
    \textbf{dens.} & 0.74 & 0.85 & 0.94 & 0.89 & 0.91 \\
    \hline
    \textbf{hmm} & 0.74 & 0.85 & 0.94 & 0.88 & 0.91 \\
    \hline
    \textbf{poiss.} & 0.71 & 0.84 & 0.92 & 0.88 & 0.90 \\
    \hline
    \hline
    \textbf{comb.} & 0.77 & 0.87 & 0.97 & 0.89 & \textbf{0.93} \\
    \hline
  \end{tabular}
\end{table}

\begin{table}[h]
  \caption{\label{table_res_dev_asr} {\it Results obtained for the automatic transcriptions of the \textsc{dev} set with the \textsc{svm}, the four systems and their linear combination (confidence interval = 0.03)}}
  \vspace{2mm} \centering
  \begin{tabular}{|c|c|c|c|c|c|}
    \hline
    \textsc{dev} & \multicolumn{5}{|c|}{\textsc{asr}} \\
    \cline{2-6}
    & $acc_1$ & $acc_2$ & \textbf{prec.} & \textbf{rec.} & \textbf{F-sc.} \\
    \hline
    \hline
    \textbf{svm.} & 0.57 & 0.75 & 0.85 & 0.85 & 0.85 \\
    \hline
    \hline
    \textbf{cos.} & 0.68 & 0.81 & 0.91 & 0.86 & 0.89 \\
    \hline
    \textbf{dens.} & 0.68 & 0.81 & 0.90 & 0.85 & 0.87 \\
    \hline
    \textbf{hmm} & 0.65 & 0.78 & 0.86 & 0.84 & 0.85 \\
    \hline
    \textbf{poiss.} & 0.66 & 0.80 & 0.90 & 0.85 & 0.87 \\
    \hline
    \hline
    \textbf{comb.} & 0.69 & 0.82 & 0.92 & 0.86 & \textbf{0.89} \\
    \hline
  \end{tabular}
\end{table}

\begin{table}[H]
  \caption{\label{table_res_test_man}{\it Results obtained for the manual transcriptions of the \textsc{test} set with the \textsc{svm}, the four systems and their linear combination (confidence interval = 0.03)}}
  \vspace{2mm} \centering
  \begin{tabular}{|c|c|c|c|c|c|}
    \hline
    \textsc{test} & \multicolumn{5}{|c|}{\textsc{man}} \\
    \cline{2-6}
    & $acc_1$ & $acc_2$ & \textbf{prec.} & \textbf{recall} & \textbf{F-sc.} \\
    \hline
    \hline
    \textbf{svm.} & 0.60 & 0.75 & 0.85 & 0.83 & 0.84 \\
    \hline
    \hline
    \textbf{cos.} & 0.67 & 0.81 & 0.90 & 0.85 & 0.88 \\
    \hline
    \textbf{dens.} & 0.64 & 0.78 & 0.87 & 0.84 & 0.85 \\
    \hline
    \textbf{hmm} & 0.66 & 0.80 & 0.89 & 0.87 & 0.88 \\
    \hline
    \textbf{poiss.} & 0.62 & 0.77 & 0.85 & 0.85 & 0.85 \\
    \hline
    \hline
    \textbf{comb.} & 0.66 & 0.81 & 0.89 & 0.87 & \textbf{0.88} \\
    \hline
  \end{tabular}
\end{table}

\begin{table}[H]
  \caption{\label{table_res_test_asr}{\it Results obtained for the automatic transcriptions of the \textsc{test} set with the \textsc{svm}, the four systems and their linear combination (confidence interval = 0.03)}}
  \vspace{2mm} \centering
  \begin{tabular}{|c|c|c|c|c|c|}
    \hline
    \textsc{test} & \multicolumn{5}{|c|}{\textsc{asr}} \\
    \cline{2-6}
    & $acc_1$ & $acc_2$ & \textbf{prec.} & \textbf{recall} & \textbf{F-sc.} \\
    \hline
    \hline
    \textbf{svm.} & 0.52 & 0.71 & 0.80 & 0.83 & 0.81 \\
    \hline
    \hline
    \textbf{cos.} & 0.63 & 0.76 & 0.86 & 0.80 & \textbf{0.83} \\
    \hline
    \textbf{dens.} & 0.55 & 0.71 & 0.81 & 0.78 & 0.79 \\
    \hline
    \textbf{hmm} & 0.55 & 0.71 & 0.79 & 0.78 & 0.79 \\
    \hline
    \textbf{poiss.} & 0.59 & 0.74 & 0.83 & 0.79 & 0.81 \\
    \hline
    \hline
    \textbf{comb.} & 0.57 & 0.72 & 0.81 & 0.78 & 0.80 \\
    \hline
  \end{tabular}
\end{table}

\subsection{Analysis of the performance of each proposed method and the linear combination of their results}

\label{subsec:res_analysis}

The results reported in Tables~\ref{table_res_dev_man}, \ref{table_res_dev_asr}, \ref{table_res_test_man} and \ref{table_res_test_asr} show that the results obtained with the proposed classification methods are not inferior and in many cases are significantly better than those obtained with \textsc{svm}. Furthermore, they also show that just a linear combination of the classifier outputs does not provide significant improvements over the best results obtained with one of the four proposed classifiers.

Some discrepancies between the results obtained with the \textsc{dev} and the \textsc{test} sets can be explained by the fact that parts of the \textsc{test} data were collected during the summer while all the \textsc{train} and \textsc{dev} data were collected in other seasons. Such a consideration is important since during the summer the proportion of traveling visitors with specific problems, speaking styles and accents is significantly higher than in the other seasons. Based on these considerations,  \textsc{lm} probabilities are expected to exhibit some variation in the summer period. In order to evaluate this expectation, samples of the same time period have been analyzed in a sequence corresponding to date and time of their collection. A set of adaptation data has been collected using words of the 1-best sequence of word hypotheses generated by the \textsc{asr} component. Using these data for \textsc{lm} adaptation resulted in minor reductions of the \textsc{wer} and minor theme classification improvements that are not worth reporting here, over representing contexts with statistics of words that coexist in a conversation.

The possibility of modeling sequences of theme mentions in a multi-theme conversation has been also considered and the manual transcriptions of the entire train set have been analyzed for this purpose. Unfortunately no reliable cues have been identified to model a theme switch by the customer. Nevertheless, such an analysis suggested that hypothesization of concepts about specific theme contents may benefit from the knowledge of some extent of the concept mention. Such an investigation is part of future work.

A careful analysis has also been made for errors due to discrepancies between a correct theme hypothesization from manual (\textsc{man}) and automatic (\textsc{asr}) annotations. It was observed that most of these errors concern the hypothesization of the \textsc{trsh} class that is not a theme but an indication that the conversation content does not concern the application domain. Errors may concern the hypothesization of \textsc{trsh}. This is the cause of 40\% of the errors in \textsc{dev} and 36,59\% of the errors in \textsc{test}. The main reason for this is the large variability of the \textsc{trsh} lexicon that makes it difficult to model and classify it as a theme.  A model and a classifier that are suitable for theme classification may not work well for rejecting out of domain conversations and different approaches should probably be investigated to further improve theme classification even if the results reported in this paper are more than satisfactory for a complete real-life application.

\section{Decision strategy}

\label{sec:strat}

The design of the decision strategy is inspired by the application requirements. The purpose is to obtain, with little or no human effort, a survey with enough correctly annotated samples for estimating, with sufficient confidence, proportions of user problem themes in a given time period.  For this purpose, a large enough and accurately annotated portion of a specific corpus (\textsc{dev}, \textsc{test}) has to be obtained from the entire corpus. Such a selection is composed with a sequential decision strategy conceived to progressively augment the survey with samples that cause an acceptable accuracy decrease. The strategy is represented by a tree. A predicate having a conversation $y$ and a set is associated with each node of the tree. Node predicates describe binary relations having $\mathbf{pred}(y,rs)$ where $\mathbf{pred}$ represents the relation type and $rs$ is a variable that takes values in sets of conversations. For the predicate associated with the root, values of $rs$ can be the \textsc{dev} set for validating the strategy or the \textsc{test} set for evaluating it. A predicate associated with a tree node returns $\mathbf{true}$ after evaluating a specific function on specific properties extracted from specific bindings of its arguments.

An initial version of the strategy is constructed as described in the following subsection. For this strategy, the truth of the node predicates is evaluated by asking a question formulated in terms of agreement among the classifications of the four different systems introduced in section \ref{sec:archi}. The motivation is that when the same result is obtained by the four systems using diffuse and localized features the consensus should correspond to a reliable decision. Furthermore, confidence of partial consensus based decisions is expected to decrease with the level of consensus.  

\subsection{Initial consensus strategy}

\label{ssec:init_strat}

Three predicates about full and partial consensus among the four systems are associated to consensus sets obtained with the initial version of the strategy tree. Each predicate has the form $\mathbf{MAJ_m}(y, rs) \ (m=4,3,2)$.

Predicate $\mathbf{MAJ_4}(y,rs)$ is associated with the root of the strategy tree and is evaluated to $\mathbf{true}$ when the four systems hypothesize the same themes for conversation $y$ in set $rs$. The entire development set (\textsc{dev}) is initially used as value of $rs$ for a preliminary assessment of the conceived strategy. Let $\{\mathbf{true}, \mathbf{false}\}$ be the possible values of $\mathbf{MAJ_4}(y,rs)$. Based on these values, the following subsets of $rs$ are formed:

\begin{equation}
  YRQ_1 := \{y \in rs \ | \ \mathbf{MAJ_4}(y,rs) = \mathbf{true} \}
  \label{eqn1}
\end{equation}

\begin{equation}
  NRQ_1 := \{y \in rs \ | \ \mathbf{MAJ_4}(y,rs) = \mathbf{false} \}
  \label{eqn2}
\end{equation}

Sets $YRQ_1$ and $NRQ_1$ are subsets of the value of variable $rs$. This value indicates the set associated with the mother node of $YRQ_1$ and $NRQ_1$ in the decision tree. Associations of this type can be inferred from the decision tree structure and will not be mentioned in the following for the sake of simplicity.
 
A decision consisting in making a survey only with conversations in  $YRQ_1$ is evaluated by two parameters, namely \textit{accuracy} ($acc_1$), as defined in equation~\ref{eq:ACC1}, and \textit{coverage}, computed as follows:

\begin{equation} 
  cov(YRQ_1) = \frac{|YRQ_1|}{|rs|}
  \label{eqn7}
\end{equation}

Let $\text{set}(\mathbf{MAJ_4}) := YRQ_1$,  $\text{size}(\text{set}(\mathbf{MAJ_4})) := |YRQ_1|$ and let

\begin{equation} 
   corr(YRQ_1) =  \sum_{y \in YRQ_1}  g(\gamma(y), M(y))
\end{equation}

where $g(\gamma(y), M(y))$ has the same meaning as in equation~\ref{eq:ACCAUX}.

\medskip

For $rs := \textsc{dev}$, the following values have been observed:

\begin{itemize}

  \item $\text{size}(\text{set}(\mathbf{MAJ_4})) = 126, \ corr(\text{set}(\mathbf{MAJ_4}))=116$

  \item $\text{size}(NRQ_1) = 70, \ \text{size}(\textsc{dev}) = 196$

\end{itemize}

from which the following evaluation of using $YRQ_1$ as survey is obtained:

\begin{itemize}

\item $cov(\text{set}(\mathbf{MAJ_4})) = 0.64$

\item $acc_1(\text{set}(\mathbf{MAJ_4})) = 0.92$

\end{itemize}

Based on the above results, motivations for starting the strategy with predicate $\mathbf{MAJ_4}(y,rs)$ are listed in the following:

\begin{enumerate}

\item Coverage and accuracy are pretty high and more suitable than what could be achieved by playing with thresholds on system combinations as combination weights could not be accurately estimated with a small development set.

\item All but one conversations in $NRQ_1$ contain only one correct theme hypothesis instead of two, missing a second semantically coherent theme hypothesis. For example, a conversation is automatically annotated with only \textit{itinerary} while it was manually annotated with \textit{itinerary} and \textit{time schedule}.

\end{enumerate}

Predicates $\mathbf{MAJ_3}$ and $\mathbf{MAJ_2}$ to represent consensus of respectively three or two systems are introduced and the corresponding sets are evaluated as described in \ref{sec:init_cons_strategy}.

The results obtained with the initial strategy, considering four possible surveys made respectively with sets based on $\mathbf{MAJ_4}, \ \mathbf{MAJ_3}, \ \mathbf{MAJ_2}$ are reported in Table~\ref{table_res_dev} for $rs := \textsc{dev}$ with details of coverage, total accuracy ($acc_1$), partial accuracy ($acc_2$), precision, total recall, and F-measure.

\begin{table}[h]
  \caption{\label{table_res_dev} {\it Results of the simple consensus strategy on the automatic transcriptions of the \textsc{dev} set (confidence interval = 0.03)}}
  \vspace{2mm}
  \centering
  \begin{tabular}{|c|c|c|c|c|c|c|}
    \hline
    \textsc{dev} & \multicolumn{6}{|c|}{\textsc{asr}} \\
    \cline{2-7}
    &\textbf{cov.} & $acc_1$ & $acc_2$ & \textbf{prec.} & \textbf{rec.} & \textbf{F-sc.} \\
    \hline
    $\text{set}(\mathbf{MAJ_4})$ & 0.64 & 0.92 & 0.96 & 0.99 & 0.62 & 0.76 \\
    \hline
    $\text{set}(\mathbf{MAJ_3})$ & 0.86 & 0.87 & 0.93 & 0.98 & 0.80 & 0.88 \\
    \hline
    $\text{set}(\mathbf{MAJ_2})$ & 1 & 0.81 & 0.88 & 0.94 & 0.89 & 0.91 \\
    \hline
  \end{tabular}
\end{table}

An analysis of the errors in the \textsc{dev} set shows that many of them can be clustered and annotated for further processing. This possibility was investigated leading to an augmentation of the decision tree as described in the next subsection.

\subsection{Strategy refinements}

The errors observed with the application of the initial strategy are due to limits of the classifiers, the types of features used, out of vocabulary words (\textsc{oov}), the errors in the 1-best sequences of words hypotheses generated by the \textsc{asr} system and the simplicity of the strategy itself.

It appears reasonable and interesting to refine the strategy with the purpose of recovering some of the above mentioned errors. The recovery strategy adds nodes and branches to the initial decision strategy tree and executes additional decision processes specific for each of the consensus sets. Each process is conceived after a human analysis of error types in each of the above sets in the \textsc{train} and \textsc{dev} sets. For the sake of clarity, in the following a conversation will be indicated by $y_i$ and a theme will be indicated by adding an index to the variable $t$.

The considered types of possible errors are listed below.

\begin{itemize}

\item DT: the deletion of a theme $t_b$ in a multiple theme conversation manually annotated with $t_a t_b$ for which only $t_a$ has been automatically annotated. Notice that the order of theme hypotheses is not relevant for the application task even if it could be inferred using the results of \textsc{density} and \textsc{hmm}. 

\item ST: the substitution of a theme $t_c$ with a theme $t_d$.

\item IT: the insertion of a theme $t_i$ in a conversation manually annotated with $t_a$ and automatically annotated with $t_a t_i$.

\end{itemize}

The observation of the errors in the \textsc{train} and \textsc{dev} sets shows that most of them are of the type DT, and the errors of the type ST and IT appear in a limited number of specific conditions, the most frequent of which concerns annotations with \textit{trash} (\textsc{trsh}), a class that is difficult to characterize because of the variability of its semantic content.



The errors in the DT cases automatically annotated with \textsc{trsh} appear to be mostly due to \textsc{asr} errors or to words such as street names and lost objects whose mention is not present in the \textsc{asr} vocabulary. Other frequent errors are due to the detection of a fare without an indication to what item it applies to. This is explained by the failure to hypothesize a type of transportation card, whose mention is present in the lattice of word hypotheses but not in the 1-best sequence.

The error analysis of the \textsc{train} and \textsc{dev} sets reveals that the just mentioned types of errors appear for theme hypotheses specific to each consensus set. It is thus reasonable to perform recovery with a small set of precondition-action rules in which a precondition is a logical expression of hypotheses generated by the four systems and appearing in a specific consensus set. The consequent action consists in performing insertions and/or deletions of theme hypotheses if the modifications are supported by the evidence of new, theme specific features evaluated with the lattice of word hypotheses. Furthermore, the actions must be coherent with the application ontology and the precondition content.

The introduction of a specific set of features for each theme is an important novelty of the proposed approach. These thematic feature sets have a limited size and are obtained automatically from the lattices of all the \textsc{train} samples. Features for each theme are selected based on their purity in the \textsc{train} set. Features are words, word classes and bigrams with a variable distance between the mention of their constituents in a time window that may involve up to three conversation turns. These distance bigrams appear to have high purity expressing distance relations between facts and actions that characterize each theme.

In this way, it is expected that a theme hypothesis receives a high score when the features characterizing it have high evidence in contrast with features characterizing other themes. Features are scored with their contribution to reduction of equivocation (\textsc{re}) when they are used to hypothesize the theme they belong to.

Relevant facts and complementary information pertinent for each theme are concepts listed in Table~\ref{theme_concepts}. The expression of these concepts is assumed to be made of words and (distant) bigrams.

\begin{table}[h!]
  \caption{\label{theme_concepts}{\it Theme related concepts expressed with selected words, bigrams and coexistence of them in a dialogue turn.}}
  \vspace{2mm} \centering
  \begin{tabular}{|l|l|l|}
    \hline
    \textbf{Traffic state (ts)} & \textbf{Time schedule} & \textbf{Itinerary} \\
    \hline
    - ts inquiry/response & - schedule mention & - itinerary reference \\
    - ts causal facts & - start hours minutes & - itinerary start \\
    - tf situation & - period reference & - itinerary end \\
    - tf state timing & & - itinerary connections \\
    \hline
    \hline
    \textbf{Cards} & \textbf{Fares} & \textbf{Lost and found (lf)} \\
    \hline
    - specific card mention & - products & - lf inquiry/response \\
    \textsc{nvgo} and \textsc{vgc} & - suburban extension & - lf objects \\
    - card status & - price & - lf actions \\
    - card related actions & - payment modality & - lf addresses\\
    - required documents & - restrictions & \\
    - accounting & & \\
    \hline
    \hline
    \textbf{Contravention} & \textbf{Special offers (so)} & \textbf{Complementary} \\
    & & \textbf{information} \\
    \hline
    - contravention mention & - special offer mention & - transportation mean \\
    - status & - so product & - generic problem \\
    - payment & - so price &  mention \\
    - procedure & - conditions & - call transfer to \\
    - complain & & specific services \\
    - declare usurpation & & - specific addresses and \\
    - request indulgence & & telephone numbers \\
    - motivation for request & & - qualitative reference \\
    - legal aspects & & to time \\
    & & - relation with other \\
    & & city entities \\
    \hline
  \end{tabular}
\end{table}

The expressions of each concept are listed in a record of features associated with the concept. Concepts such as time and price are decomposed into constituents. Each constituent is represented by a word or a bigram feature.

In some cases the detection of only part of the constituents of a concept may be useful for recovering a deletion. In other cases it may introduce a false alarm. For this reason, the use of lattice features has been constrained with specific preconditions expressing the detection of a significant but incomplete amount of theme features.

Other types of features could be considered but have not been used. For example, prosodic features may contribute to characterize complaints or requests of indulgence. The main reason is that it is difficult to extract reliable prosodic features in the customer turns because the acoustic environments may be affected by a large variety of noise types and intensity.

In order to introduce the form of preconditions let us define:

\begin{itemize}

\item $H_c(i, cs)$: the automatic annotation by system \textsc{cosine} of conversation $y_i$ in the consensus set $cs$.

\item $H_d(i, cs)$: the automatic annotation by system \textsc{density} of conversation $y_i$ in the consensus set $cs$.

\item $H_h(i, cs)$: the automatic annotation by system \textsc{hmm} of conversation $y_i$ in the consensus set $cs$.
  
\item $H_P(i, cs)$: the automatic annotation by system \textsc{poisson} of conversation $y_i$ in the consensus set $cs$.

\end{itemize}

Let $(mH \rightarrow A), \ (m = 2, 3, 4)$ represent the fact that $m$ and only $m$ annotations of the four systems are equal to $A$. For example, let assume that, for a conversation in the \textsc{dev}, it has been observed $H_d(i,cs) = H_h(i,cs)  = H_P(i,cs)  = \textsc{objt}$ (\textit{lost\_and\_found} theme) and $H_c(i, cs) = \textsc{objt}\_\textsc{nvgo}$ (lost a \textsc{nvgo} transportation card). 

This situation is generalized leading to the following precondition representation:

\[
P3(L,Q): (3H \rightarrow L) \wedge  (1H \rightarrow LQ)
\]

Five general forms have been introduced, one for $YRQ_1$ (see Subsection~\ref{ssec:init_strat}), one for $YRQ_2$ (see \ref{sec:init_cons_strategy}) and three for $NRQ_2$ (see \ref{sec:init_cons_strategy}).

A general form can be specialized by binding the variables $L$ and $Q$ to specific values as in the previous example:

\[
L = \textsc{objt} \text{ and } Q = \textsc{objt}\_\textsc{nvgo}
\]

Specialized preconditions have been manually derived from the semantic coherence relations introduced later on after the actions.

An action is introduced as consequence of each specialized precondition.

An action $AC_{j(n)}$ asserting an annotation modification $H_i{j(n)}$ of a theme annotation for conversation $y_i$ includes two processes, namely:

\begin{itemize}

\item assert a semantically coherent modification $H_{i}{j(n)}$,

\item evaluate the evidence of $H_{i}{j(n)}$.

\end{itemize}

A precondition-action rule is written as follows:

\[
PC_{j(n)}\Longrightarrow AC_{j(n)}
\]

here $j(n)$ indicates the $j$-th specification of the $n$-th general precondition. Symbol $\Longrightarrow$ links a precondition to its corresponding action.

Action $AC_{j(n)}$ is formally represented as follows:

\[
AC_{j(n)}: \text{assert}(H{j(n)}) \wedge \text{evaluate}(H{j(n)})
\]

The process $\text{assert}(H{j(n)})$ generates hypotheses coherent with the application ontology according to the content of a look-up table corresponding to precondition $j(n)$. The application ontology has been inferred from the application documentation and the look-up table has been compiled based on the analysis of the errors of the \textsc{train} set. Both activities have been performed by a human expert. Common to many application domains is the use semantic knowledge containing structures representing a theme with its related facts and some complementary information. These structures are represented by lambda-expressions of this type:

\begin{equation}
  (\lambda \ (x \ y \ z) \ (\text{theme}(x) \wedge \text{facts}(x, y) \wedge \mathrm{compl\_inf_1}(x, y, z)))
\end{equation}

The semantic content of a conversation is obtained by binding the variables $x, y, z$ to specific values. Notice that a value can be another semantic structure with all variables bound to specific values. For example, if $x = \textit{traffic\_state}$, then $y$ may be bound to the elements of the set $ts:= \{\text{strike}, \text{delay}, \text{accident}, \text{anomaly}\}$. Variable $z$ can be bound by other complementary information such as \textit{time} and \textit{location}. In many application domains, there are specific facts for each theme and specific relations between different themes.

Given the type of application in which conversations have to follow a defined protocol, it is reasonable to assume that co-existence of mentions of fact predicates and coherent argument values are sufficient conditions for expressing themes characterized by these predicates.

The conversation protocol implies that only themes with semantically coherent facts may coexist in a conversation. This may not be always the case in practice. Nevertheless, this type of semantic coherence can be used for imposing constraints for recovering hypothesized theme errors. For example, a mention of the theme \textsc{objt} may share the fact that the lost object is a type of \textsc{nvgo} card, thus the two themes may coexist. In order to further constrain the application of the recovery strategy, frequent errors observed in the \textsc{train} and \textsc{dev} sets have been clustered into types. The types of errors listed below have been used in the introduced coherent recovery actions:

\begin{itemize}

\item \textit{fare} in conjunction with \textit{itinerary} and \textit{transportation cards}.

\item \textit{time} in conjunction with \textit{itinerary} and \textit{traffic\_state}.

\item \textit{lost\_found} in conjunction with \textit{cards}.

\item \textit{itinerary} in conjunction with \textit{traffic\_state, cards, lost\_found}.

\item \textit{fine} in conjunction with \textit{cards} and \textit{loss}.

\end{itemize}

Refinement strategy actions are executed only if specific conditions and the asserted hypotheses ($H{j(n)}$) are verified.

The procedure for hypothesis evaluation is now described.

A feature set $\Phi_k := \{\varphi_1^k,....,\varphi_n^k,.....,\varphi^k_{N_\Phi}\}, \ k \in \{1, ..., K\}$ of size $N_\Phi$ is defined for each theme $t_k$. Only conversations in the train set annotated with a single theme are considered and used for inferring features of $\Phi_k$ for each $k$. The $i$-th conversation of the entire corpus is described by the acoustic features $A_i$. The mutual information between the annotated theme $t_k$ and $A_i$ is defined as:

\begin{equation}
  I(t_k,A_i)=H(t_k)-H(t_k|A_i)
\end{equation}

where $H(t_k)$ is the entropy of theme $t_k$ and $H(t_k|A_i)$ is the conditional entropy of theme $t_k$ computed with features $\Phi_k$ extracted from $A_i$.

Details on the computation of $H(t_k|A_i)$ are given in \ref{sec:rec_strat}.

The automatic feature selection process starts by considering a word vocabulary $V_1$ obtained from the application vocabulary $V$ (5,782 words) after removal of the words of a stop list. A list $LB$ of distant bigrams have been added to $V_1$ by considering the association of a word in $V_1$ with any word in V in a window spanning three turns. The elements of $V_1$, $LB$ and a small set of abstract classes such as time and price are considered as possible features belonging to an initial set $\Phi_0$. Set $\Phi^k$ is formed by computing, for every element $\varphi_x\in\Phi_0$ the purity measured by the probability $\mathbb{P}(t_k |\varphi_x)$, whose computation is described in \ref{sec:rec_strat}.

The size of the feature set has been obtained by automatically annotating the conversations in the train set manually annotated with only one theme. Automatic annotation was based on the following decision rule:

\begin{equation}
  \widehat{T} = \argmax_j \ H(t_j|A_i)
\end{equation}

After ordering by feature purity the feature set of each theme the same size of feature set candidates was considered for all themes. The size was progressively increased from 50 to 1,000 by steps of 5. For each step, the accuracy of theme hypothesization was computed in the \textsc{train} set. The process of growing the feature set size stopped when no tangible improvements in the classification accuracy was observed. A value $N_\Phi = 420$ was found in this way.

Overall, 1,440 word features and 1,657 distant bigram features were obtained. The highest word purity for each theme varies between 0.6 and 0.91 depending on the theme, while the lowest word purity varies between 0.12 and 0.4.

Even with these features, some combinations of theme hypotheses were not well characterized. For example, it was not possible to have an exhaustive list of lost objects. For this purpose, it was considered useful to infer \textit{complete topic mentions}. \textit{Complete topic mentions} were manually compiled by selecting, abstracting and combining into patterns some automatically obtained features. For example, pattern structures were derived for abstractions of mentions of departure, arrival, connection to describe an itinerary. Patterns characterizing actions describing loss of an object without specifying the object type were also compiled into an abstract class. Some minor improvements were observed after the introduction of \textit{complete topic mentions} features.

As an example of how recovery actions are applied, let us consider the case of recovering the deletion of a theme $AA$ when theme $B$ has been hypothesized. The following decision rule is applied:

\[
\left.
\begin{aligned}
  & \mathbf{semantically\_coherent}(AA, B) & \wedge \\
  & \mathbf{greater\_than}(\text{score}(B), \text{TH}(B)) & \wedge  \\
  & \big [ \mathbf{rank\_RE\_FIRST}(B) \ \vee \\
  & \ (\mathbf{rank\_RE\_FIRST}(AA) \ \wedge \\ 
  & \ \ \mathbf{rank\_RE\_SECOND(B)}) \big ] \\
\end{aligned}
\right \}
\Longrightarrow \text{assert}(AA\_B)
\]


Scores are computed with the reduction of equivocation (\textsc{re}) process.
These scores are obtained with features of a conversation whose mention is omitted in the above relation for the sake of simplicity. $\text{TH}(B)$ is a threshold evaluated from the \textsc{dev} set to optimize separation between positive and negative examples. With a larger \textsc{dev} set, $\text{TH}(B)$  could be estimated as the value that minimizes the relative entropy between manually and automatically annotated distributions of theme proportions.  This threshold is applied to all consensus sets. Predicates $\mathbf{rank\_RE\_FIRST}(B)$ and $\mathbf{rank\_RE\_SECOND}(B)$ are evaluated to \textbf{true} when theme $B$ is respectively scored first or second by the \textsc{re} process. Decision of removing false insertion themes is applied in specific conditions when the theme position in the \textsc{re} ranking is not among the first three.

For specific conditions, these types of decisions were also considered:

\begin{itemize}

\item composition of themes hypothesized by different systems,

\item recovery of a theme from \textsc{trsh}.

\end{itemize}

Composition decisions were made based on positions and scores of the themes to be combined in the following situations:

\begin{itemize}

\item highest \textsc{re} score for single insertion or \textsc{trsh} recovery,

\item first two positions and co-presence in the hypotheses for the composition of two themes,

\item composition of three themes in case of co-presence in different system hypotheses of only three semantically coherent themes and high \textsc{re} scores for them.

\end{itemize}

Based on the frequency of the observed errors in the \textsc{dev} and \textsc{train}, recovery for data in $\text{set}(\mathbf{MAJ_4})$ was considered only for the following semantic relations:

\begin{itemize}
  
\item $B = \mathit{fare}$ in conjunction with $AA =\textsc{nvgo}$ (transportation card),

\item $B =\mathit{time} \text{ or } B = \textsc{nvgo}$ in conjunction with $AA = \mathit{itinerary}$,

\item recovery of $B = \mathit{fine} \text{ from } AA = \mathit{trash}$.

\end{itemize}

The same cases as for $\text{set}(\mathbf{MAJ_4})$  were considered for $YRQ_2$ with the addition of

\begin{itemize}
 
\item $B = \mathit{traffic\_state}$ in conjunction with $AA = \mathit{itinerary}$,

\item $B = \mathit{lost}$ in conjunction with $AA = \textsc{nvgo}$,

\item recovery of $B = \textsc{vgc} \text{ card} \text{ or } B = \mathit{traffic\_state} \text{ from } AA = \mathit{trash}$.

\end{itemize}

Recovery in $cs_3$ (subset defined in \ref{sec:init_cons_strategy}) is more difficult since possible major problems may be caused by errors in the 1-best sequence or by limits of some of the four systems. When the hypotheses generated by the four systems are all different it is likely that the 1-best sequence contains too many errors. In both cases \textsc{re} ranking and scores are used for recovering deletions and also for obtaining a new majority vote with the addition of the \textsc{re} ranking.

The results obtained with the initial strategy on the test set are reported in Table~\ref{table_res_cons_test} and those obtained with the  addition of the recovery strategy are reported in Table~\ref{table_res_recovery_test}.

\begin{table}[h]
  \caption{\label{table_res_cons_test} {\it Results of the simple consensus strategy on the automatic transcriptions of the \textsc{test} set (confidence interval = 0.03)}}
  \vspace{2mm}
  \centering
  \begin{tabular}{|c|c|c|c|c|c|c|}
    \hline
    \textsc{test} & \multicolumn{6}{|c|}{\textsc{asr}} \\
    \cline{2-7}
    &\textbf{cov.} & $acc_1$ & $acc_2$ & \textbf{prec.} & \textbf{rec.} & \textbf{F-sc.} \\
    \hline
    $\text{set}(\mathbf{MAJ_4})$ & 0.54 & 0.88 & 0.92 & 0.96 & 0.50 & 0.66 \\
    \hline
    $\text{set}(\mathbf{MAJ_3})$ & 0.80 & 0.81 & 0.87 & 0.92 & 0.70 & 0.80 \\
    \hline
    $\text{set}(\mathbf{MAJ_2})$ & 0.98 & 0.74 & 0.82 & 0.87 & 0.82 & 0.84 \\
    \hline
  \end{tabular}
\end{table}

\begin{table}[h]
  \caption{\label{table_res_recovery_test} {\it Results of the recovery  strategy on the automatic transcriptions of the \textsc{test} set (confidence interval = 0.03)}}
  \vspace{2mm} \centering
  \begin{tabular}{|c|c|c|}
    \hline
    \textsc{test} & \multicolumn{2}{|c|}{\textsc{asr}} \\
    \cline{2-3}
    & \textbf{cov.} & $acc_1$ \\
    \hline
    set($\mathbf{MAJ_4}$) & 0.54 & 0.89 \\
    \hline
    set($\mathbf{MAJ_3}$) & 0.80 & 0.86 \\ 
    \hline
    set($\mathbf{MAJ_2}$) & 0.98 & 0.80 \\
    \hline
    \textbf{All test} & 1 & 0.80 \\
    \hline
  \end{tabular}
\end{table}

Results show a statistically significant contribution of the recovery strategy on $acc_1$ for high coverage with respect to the use of just the initial consensus strategy.

The numbers of concept instances annotated in the manual transcriptions (indicated in the following as \textsc{man}), and detected from the word hypotheses generated by the \textsc{asr} (most likely sequence and lattice) for the \textsc{dev} (8,646 turns~--~196 dialogues) and the \textsc{test} (27,497 turns~--~578 dialogues) sets are reported in Table~\ref{table_concepts}.

\begin{table}[h]
  \caption{\label{table_concepts} {\it Number of concept instances for the \textsc{dev} and \textsc{test} sets}}
  \vspace{2mm} \centering
  \begin{tabular}{|c|c|c|c|}
    \hline
    & \textsc{man} & \textsc{asr} one-best & \textsc{asr} lattice \\
    \hline
    \textsc{dev} & 26,256 & 18,906 & 21,536 \\
    \hline
    \textsc{test} & 80,764 & 47,725 & 59,166 \\ 
    \hline
  \end{tabular}
\end{table}

The increase of relevant concepts detected in the lattice with respect to the one best sequence is significant. This is done at the expenses of 30\% insertions that have little impact on theme recovery since most of them are filtered out by the selective recovery procedure.

By observing the concept mentions in the manually annotated concepts it appears that errors in automatic concept hypothesization are essentially due to \textsc{asr} errors.

Published experimental results on the \textsc{decoda} corpus refer to the identification of the dominant theme. Among them, the highest accuracy of 85\% is reported in \citep{morchid2014vector}.

This value has been obtained with a subset of the corpus used in this paper annotated with a single theme label and a large number of \textsc{lda} feature sets extracted in hidden spaces of different size. This single theme is annotated with one of 8 theme labels. This label set is obtained from the one used in the experiments of this paper by removing \textsc{trsh} and \textsc{vgc}. Such a classification is useful in practice if out domain conversations are discarded by another method and proportions are evaluated for problems of transportation cards ignoring the card type.

Considering that the confidence interval of the results is 0.03, the accuracy $acc_1$ reported in this paper for multiple theme labelling is slightly inferior to the best accuracy reported for just one theme labelling with a smaller set of labels and a smaller corpus. It is worth mentioning that the results reported in this paper for multiple theme are superior or inside the confidence interval when compared with results reported in \citep{morchid2014improving} for the identification of one of the above mentioned 8 theme labels obtained with Gaussian and \textsc{svm} classifiers using different types of features including \textsc{lda} hidden topic features computed in a single hidden space.

\subsection{Error analysis}

The errors observed in the \textsc{test} set of $YRQ_1$ using the initial strategy were 45 in 321 conversations. Among them, 6 deletions were corrected to $X\_Y$ when $X$ was hypothesized. Using only the complete mentions, 4 more errors of the same type were corrected while 1 false insertion was generated. Among the remaining errors, there were 15 deletion errors of $Y$ with $X$ correct and two insertion errors with $X$ correct. The other errors where 9 substitutions involving \textit{trash} indicating difficulties in characterizing this type of rejection and the others were confusions between \textit{itinerary} and \textit{traffic\_state} and between \textit{lost\_and\_found} and \textit{cards}. Listening to these conversations, it appeared that the two themes were mentioned, but the annotators considered one of them not sufficiently relevant.

The errors observed in the \textsc{test} set of $YRQ_2$ were 44 in 131 conversations. Among them, 10 deletions were corrected to $X\_Y$ when $X$ was hypothesized. Using only the complete mentions 5 more errors of the same type were corrected.

The errors observed in the \textsc{test} set of $YRQ_3$ were 71 in 117 conversations. Among them, no false recoveries were observed, 5 deletions were corrected to $X\_Y$ when $X$ was hypothesized, 6 compositions were correctly executed by merging hypotheses from different systems, 7 recoveries from \textit{trash} were correctly performed. A minor number of corrections for $YRQ_3$ were made using the additional contribution of \textsc{re} ranking and complete mentions.

For $YRQ_2$ and $YRQ_3$, corrected  and remaining errors  were of the same nature as those in $YRQ_1$.

A large majority of the remaining errors are deletions of a theme in a conversation in which at least one of the annotated themes was correctly hypothesized. By listening to the conversations annotated with errors it appears that, in many of them, the annotators have made an intelligent choice of what was relevant, while the system ignored part of it or inserted themes that were mentioned, but were not relevant. Based on the analysis it is reasonable to conclude that, with the approach described in \citep{camelin2009error}, significant theme proportion variations can be detected in a time period only a little longer than if the manual annotations were used.

\section{Conclusion and future work}

Four systems for multiple theme hypothesization in human/human call center conversations have been considered. Two of them, \textsc{cosine} and \textsc{poisson} extract features for theme mentions in an entire conversation. Other two systems \textsc{density} and \textsc{hmm} attempt to localize theme mentions in specific zones of a conversation. A comparison of the results separately obtained with each system show little performance difference, while a study on the consensus among the results of the four systems show that there is a non-negligible proportion of conversations for which the four systems do not completely agree. Nevertheless, in such a case it is still possible to perform reliable decisions on specific conversation sets by considering majority votes. In making these decisions all four systems appear to be useful indicating that their combined use contribute to effectively process diffuse as well as localized theme mentions.

For a significant proportion of those cases for which reliable decisions cannot be made with majority vote strategy it is still possible to trigger a new process with new features for obtaining a high coverage survey with accurately annotated themes. The overall result is that the proposed approach makes it possible to produce practically useful theme proportions in spite of \textsc{asr} errors and the imprecision of classification methods.

The recovery strategy has some precondition-action rules that depend on the application ontology and other that are applicable to a large variety of application domains. This latter set contains rules for recovering false classification as \textsc{trsh} due to concept deletion errors and false insertions of concepts leading to the hypothesization of a domain theme rather than classifying the conversation as \textsc{trsh}. Another general case is the deletion of concepts whose mentions are short words that are likely to be deleted in the 1-best sequence of word hypotheses. These short words are often confused by inserting or deleting one short phoneme such as a plosive consonant. Specific recovery rules can then be conceived by observing a limited number of examples in the train set and using them for setting appropriate contexts for searching instances in the lattice of word hypotheses.

Future work should attempt to provide accurate estimations of the amount of time periods in which corpora have to be collected in order to extract practically useful surveys.

More detailed semantic information has also to be extracted under the control of the application domain ontology. New methods have to be introduced for estimating the amount and degree of user problem solution and for composing all the extracted semantic contents to produce conversation reports. These reports are not necessarily summaries. They should rather be semantic structures of strictly relevant information for the application task.

\section*{Acknowledgements}

This work has been supported by the French National Research Agency (\textsc{anr}) with Project \textsc{decoda}, contract \textsc{anr}-09-\textsc{coord}-005, and the French cluster \textsc{scs} (Secured Communicating Solutions). The corpus used for the experiments has been provided by the \textsc{ratp} (Paris public transportation system).


\newpage

\appendix

\section{Theme density: computation details}

\label{sec:dens_comp}

\subsection{Recurrence formulas}

According to equation~\eqref{eq:ED1}, the density for theme $t$ at the $i$-th position of a conversation is evaluated as follows:

\begin{equation}
  d_{t}(p_i) = \frac{\sum_{j=1}^n
    \frac{w_t(p_j)}{\lambda^{d_j}}}{\sum_{j=1}^n
    \frac{1}{\lambda^{d_j}}} \ (i = 1, ..., n)
\end{equation}

where $w_t(p_j)$ denotes the contribution to theme $t$ of the features located at the $j$-th position; $\lambda \geqslant 1$ is a parameter of sensitivity to proximity and $d_j:= |i - j|$ is the distance between positions $p_i$ and $p_j$.

The density $d_t(p_i)$ for theme $t$ at the $i$-th position can first be expressed as follows:

\begin{equation}
  \begin{aligned}
    d_{t}(p_i) & = \frac{\sum_{j=1}^n
      \frac{w_t(p_j)}{\lambda^{d_j}}}{\sum_{j=1}^n
      \frac{1}{\lambda^{d_j}}} \\
    & = \frac{\sum_{j=1}^{i-1} \frac{w_t(p_j)}{\lambda^{i-j}}
      + w_t(p_i) + \sum_{j=i+1}^n \frac{w_t(p_j)}{\lambda^{j-i}}}
    {\sum_{j=1}^{i-1} \frac{1}{\lambda^{i-j}}
      + 1 + \sum_{j=i+1}^{n} \frac{1}{\lambda^{j-i}}} \\
  \end{aligned}
  \label{eq:ED2}
\end{equation}

\bigskip

With reference to the $i$-th position, let $L(p_i) = \sum_{j=1}^{i-1} \frac{w_t(p_j)}{\lambda^{i-j}}$ be the weighted sum of the left-hand side thematic contributions and $R(p_i) = \sum_{j=i+1}^n \frac{w_t(p_j)}{\lambda^{j-i}}$ be the weighted sum of the right-hand side thematic contributions. Let $NL(p_i) = \sum_{j=1}^{i-1} \frac{1}{\lambda^{i-j}}$ be the left-hand side summation of the normalization factors and let $NR(p_i) = \sum_{j=i+1}^{n} \frac{1}{\lambda^{j-i}}$ be the corresponding right-hand side summation. Introducing these notations in equation~\eqref{eq:ED2}, one gets for $i = 1, ..., n$:

\begin{equation}
  \begin{aligned}
    d_{t}(p_i) & = \frac{L(p_i) + w_t(p_i) + R(p_i)}{NL(p_i) + 1 +
      NR(p_i)} \\
    & = \frac{L(p_i) + w_t(p_i) + R(p_i)}{NL(p_i) + 1 +
      NL(p_{n-i+1})} \\
    \label{eq:ED3}
  \end{aligned}
\end{equation}

The quantities $L(p_i), R(p_i), \text{ and } NL(p_i)$ can be computed as follows by simple recurrence on the neighbor positions:

\medskip

\textbf{Initialization:}

\begin{equation}
  L(p_1) = R(p_n) = NL(p_1) = 0
\end{equation}

\textbf{Recurrence:}

\begin{equation}
  \begin{aligned}
    L(p_i) & = \sum_{j=1}^{i-1} \frac{w_t(p_j)}{\lambda^{i-j}} \\
    & = \sum_{j=1}^{i-2} \frac{w_t(p_j)}{\lambda^{i-j}} +
    \frac{w_t(p_{i-1})}{\lambda^{i-(i-1)}} \\
    & = \sum_{j=1}^{i-2} \frac{w_t(p_j)}{\lambda^{i-1+1-j}} +
    \frac{w_t(p_{i-1})}{\lambda} \\
    & = \sum_{j=1}^{(i-1)-1} \frac{w_t(p_j)}{\lambda^{(i-1)-j} \lambda} +
    \frac{w_t(p_{i-1})}{\lambda} \\
    & = (L(p_{i-1}) + w_t(p_{i-1})) \frac{1}{\lambda} \ \ \ (\text{with } i = 2, ..., n) \\
  \end{aligned}
\end{equation}

\begin{equation}
  \begin{aligned}
    R(p_i) & = \sum_{j=i+1}^n \frac{w_t(p_j)}{\lambda^{j-i}} \\
    & = \frac{w_t(p_{i+1})}{\lambda^{(i+1)-i}} + \sum_{j=i+2}^n \frac{w_t(p_j)}{\lambda^{j-i}} \\
    & = \frac{w_t(p_{i+1})}{\lambda} +  \sum_{j=i+2}^n
    \frac{w_t(p_j)}{\lambda^{j-1+1-i}} \\
    & = \frac{w_t(p_{i+1})}{\lambda} +  \sum_{j=(i+1)+1}^n
    \frac{w_t(p_j)}{\lambda^{j-(i+1)} \lambda} \\
    & = \frac{1}{\lambda} (w_t(p_{i+1}) + R(p_{i+1})) \ \ \ (\text{with } i = (n - 1), ..., 1) \\
  \end{aligned}
\end{equation}

In a similar way, one gets:

\begin{equation}
    NL(p_i) = (NL(p_{i-1}) + 1) \frac{1}{\lambda} \ \ \ (\text{with } i = 2, ..., n)
\end{equation}

\subsection{Computation algorithm}

\label{sec:dens_algo}

The use of these recurrences for computing the thematic density is shown in the description of Algorithm~\ref{algo:dens}.

\begin{algorithm}[h]
\caption{\textsc{Theme density computation}}
\label{algo:dens}
\begin{algorithmic}[1]
  \REQUIRE $t \in \mathbb{T}, \ n \in \mathbb{N}, \ w[t, i] \in
  \mathbb{R^+}, \ \lambda \in [1; + \infty[$
  \FORALL {$t \in \mathbb{T}$}
  \STATE $L[1] \leftarrow 0$
  \STATE $R[n] \leftarrow 0$
  \STATE $NL[1] \leftarrow 0$
  \FOR {$i \leftarrow 2$ to $n$}
  \STATE $L[i] \leftarrow (L[i-1] + w[t, i-1]) \ / \ \lambda$
  \STATE $R[n-i+1] \leftarrow (R[n-i+2] + w[t, n-i+2]) \ / \ \lambda$
  \STATE $NL[i] \leftarrow (NL[i-1] + 1) \ / \ \lambda$
  \ENDFOR
  \FOR {$i \leftarrow 1$ to $n$}
  \STATE $dens[t, i] \leftarrow L[i] + w[t, i] + R[i]$
  \STATE $dens[t, i] \leftarrow dens[t, i] \ / \ (NL[i] + 1 + NL[n-i+1])$
  \ENDFOR
  \ENDFOR
\end{algorithmic}
\end{algorithm}

Positions $p_i \ (i = 1, ..., n)$ are represented by their respective indices and the density $d_t(p_i)$ of the theme $t$ at position $p_i$ is indicated as $dens[t, i]$. Similarly, the contribution $w_t(p_i)$ to the theme $t$ of the features located at position $p_i$ is indicated as $w[t, i]$.

The number of the instructions of the algorithm~\ref{algo:dens} is given by:

\begin{equation}
  |\mathbb{T}|(3 + 3(n-1) + 2n) = (5 |\mathbb{T}|) n
\end{equation}

The time complexity is thus a linear function of the number of conversation positions.

\section{Thematic skeleton of a conversation}

\label{sec:conv_skel}

\begin{figure}[h]
  \centering
    \includegraphics[width=120mm]{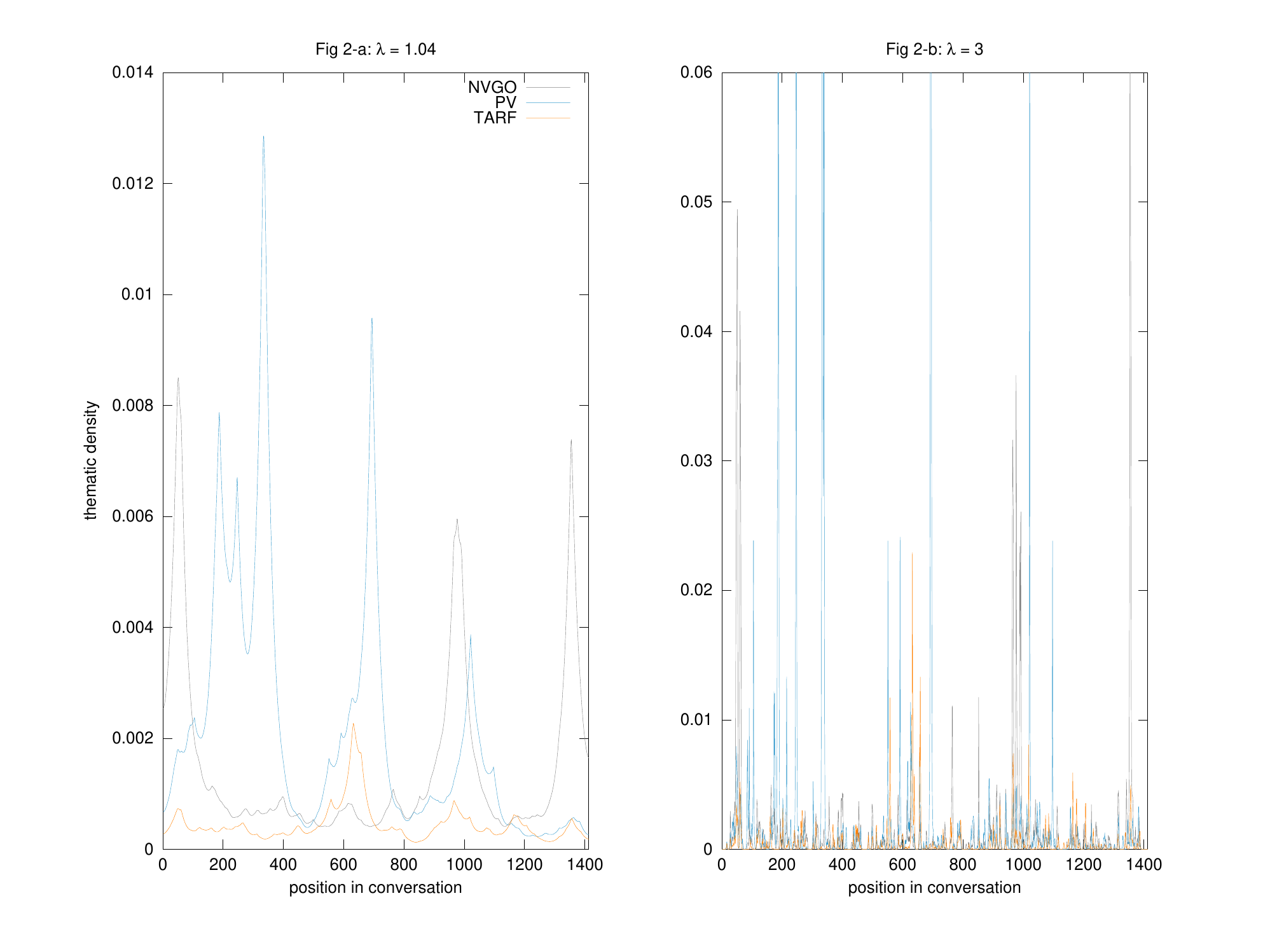}
    \caption{\label{lambda} {\it Thematic densities as function of
        position in a conversation skeleton. Densities are plotted for
        $\lambda = 1.04$ (Fig.~2-a) and $\lambda = 3$ (Fig.~2-b) and
        three themes: transportation card (indicated as
        \textsc{nvgo}), fine (indicated as \textsc{pv}) and fare
        (indicated as \textsc{tarf})}}
\end{figure}

The thematic skeleton of a conversation for three themes is shown in Figure~\ref{lambda}.

The skeletons for $\lambda = 1.04$ (Fig.~B.2-a) and $\lambda = 3$ (Fig.~B.2-b) were obtained using a manual word transcription of the conversation, which is about a fine (theme indicated as \textsc{pv}) caused by an identity theft. The customer has been fined because he used his mother's transportation card (theme indicated as \textsc{nvgo}). Two functions are plotted for the annotated themes \textit{transportation card} and \textit{fine} and a third function is plotted for the theme \textit{fare} (indicated as \textsc{tarf}) that is not annotated because not completely mentioned in the conversation and considered to be irrelevant for the task.

The figures have been obtained from a conversation whose an excerpt, as manually transcribed, is reported in the following. The conversation positions corresponding to each turn are in brackets.

\begin{itemize}
\item[--] \textit{Customer} [22--116]: I am calling because I had a problem yesterday with my \textsc{nvgo} card (...) I used my mother's card (...) The controller confiscated the card.
\item[--] (...)
\item[--] \textit{Agent} [320--338]: We are going to receive your card at the collection center.
\item[--] (...)
\item[--] \textit{Customer} [544--630]: How much the fine will be? (...) I have been told I have to pay forty euros (...)
\item[--] \textit{Agent} [642--656]: Right. It should be forty euros.
\end{itemize}

Distant contextual relations, corresponding to high values of $\lambda$, such as $3$ in Fig.~B.2-b, tend to be neglected. In these cases, incorrect decisions may be caused by isolated features that are not relevant for theme hypothesization. For example, the expressions \textit{how much} (\textit{combien} in French) in turn [544--630] and \textit{forty euros}, repeated twice in turns [544--630] and [642--656], tend to show evidence for fare with a peak of density at position 630, whereas it is just, in this case, part of a request about fine.

Local context, more appropriate for theme mention detection, is appropriately represented by a value of $\lambda = 1.04$ as in Fig.~B.2-a. This has the effect of reducing the relevance of the fare hypothesis only supported by the expressions \textit{how much} and \textit{forty euros}.

In conclusion, values of $\lambda$ such as 1.04 correspond to thematic coherence leading to more accurate decisions.

\section{Consensus strategy with three or two systems}

\label{sec:init_cons_strategy}

A predicate $\mathbf{MAJ_3}(y, NRQ_1)$ takes value $\mathbf{true}$ if there is a consensus among only three systems. According to this value, two new subsets $YRQ_2$ and $NRQ_2$ are obtained.

Considering now a new survey set, larger than $YRQ_1$ and defined as $\text{set}(\mathbf{MAJ_3}) := (YRQ_1 \cup YRQ_2)$, the following values are found:

\begin{itemize}
 
\item $\text{size}(YRQ_2) = 43, \ corr(YRQ_2) = 31$

\item $cov(\text{set}(\mathbf{MAJ_3})) = (126+43) / 196 = 0.86$

\item $acc_1(\text{set}(\mathbf{MAJ_3})) = (116+31) / (126+43) = 0.87$

\end{itemize}

A high accuracy is still obtained with a coverage much greater than that of $YRQ_1$.

With similar motivations, the predicate $\mathbf{MAJ_2}(y, NRQ_2)$, was considered and applied. It is worth considering three possible types of samples for which the predicate is asserted true. They are:

\begin{itemize}

\item $XY_3$: the automatic annotation of two systems is $X$ and the automatic annotation of the other two systems is $Y$.

\item $Za_3$: the automatic annotation of two systems is $Z$ and the automatic annotations of the other two systems are respectively $Z_1$ and $Z_2$, both different from $Z$ and such that $Z_1 \neq Z_2$.

\item $N_3$: the four systems provide four different automatic annotations.
 
\end{itemize}

Let $cs_3 \subseteq NRQ_2$ be a subset of $NRQ_2$ defined as follows:

\[
cs_3 := Za_3 \cup LC_3
\]

where $LC_3$ is the set $XY_3$  automatically annotated with the results of the linear combination of the four systems. Let:  $\text{set}(\mathbf{MAJ_2}) := (YRQ_1 \cup YRQ_2  \cup cs_3)$.

In such a case:

\begin{itemize}

\item $\text{size}(cs_3) = 26, \ corr(cs_3) = 11$

\item $cov(\text{set}(\mathbf{MAJ_2})) = 1, \ acc_1(\text{set}(\mathbf{MAJ_2}))= 0.81$

\end{itemize}

\section{Recovery strategy: computation details}

\label{sec:rec_strat}

Conditional entropy $H(t_k|A_i)$ is computed as follows:

\begin{equation}
  \begin{split}
    H(t_k|A_i) &= \\
    - \sum_{n = 1}^{N_\Phi} \left [ \mathbb{P}(t_k | \varphi_n^k, A_i) \times \mathbb{P}(\varphi_n^k | A_i) \right ] & \log_2 \left [ \mathbb{P}(t_k|\varphi_n^k, A_i) \times \mathbb{P}(\varphi_n^k | A_i) \right ] 
  \end{split}
\end{equation}

$\mathbb{P}(t_k | \varphi_n^k, A_i)$ is approximated with $\mathbb{P}(t_k | \varphi_n^k)$.

This conditional entropy can be seen as a reduction of equivocation (\textsc{re}) in the estimation of $t_k$ knowing $A_i$. The same number $N_\Phi$  of features is used for all themes.

Let $\mathbb{P}(t_k | \varphi_n^k)$ be represented as $\mathbb{P}(t_k|\varphi_x)$ for the sake of simplicity. Probability $\mathbb{P}(t_k|\varphi_x)$ is computed as follows:

\begin{equation}
  \mathbb{P}(t_k|\varphi_x) = \frac{c(\varphi_n,t_k) + K\mathbb{P}(t_k)}
  {\overset{K}{\underset{j=1}{\sum}} c(\varphi_x,t_j) + K}
\end{equation}

where $c(\varphi_x,t_k)$ is the count of the times $\varphi_x$ appears in all the conversations in the train set manually annotated with $t_j$.

Features are ranked based on their purity in each theme $t_k$ and an ordered list $L_k$ is compiled for each theme.

The presence of a candidate feature $\varphi_x^k \in L_k$ in a conversation is measured by the probability $\mathbb{P}(\varphi_x^k|A_i)$. If the feature is a distant bigram involving a word $w_m$ and its context $c_{m,m-\tau}$ the feature posterior probability is computed by the following product:

\begin{equation}
  \mathbb{P}(\varphi_x^k|A_i) = \mathbb{P}(c_{m,m-\tau}|A_i)\times \mathbb{P}(w_m|A_i)
\end{equation}

Features are considered as terms in a \textit{term frequency-inverse document frequency} score defined according to \citep{hazen2011mce}  as:

\begin{equation}
  \begin{aligned}
    \gamma_{d,k}[idf(\varphi_x^k)] 
    & = idf(\varphi_x^k) \times \underset{g\in G_n}{\sum}\mathbb{P}_g(\varphi_x^k|A_i) \\ 
    &= idf(\varphi_x^k) \times \frac
    {\underset{g\in G_{x,k}}{\sum} \lambda_g (\varphi_x^k)}
    {Z}
  \end{aligned}
\end{equation}

where $Z=\overset{K}{\underset{j=1}{\sum}} \underset{g \in G_{u,j}}{\sum} \lambda_g(\varphi_j^k)$ is the sum of the scores of all features competing with $\varphi_x^k$. $G_{u,j}$ is the set of arcs in the graph labelled with the hypothesis.




\newpage

\section*{References}

\bibliographystyle{model2-names}
\bibliography{base}

\newpage

\section*{Vitae}
\includegraphics[width=1in,clip,keepaspectratio]{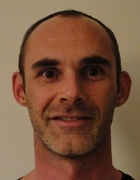}
\noindent {\bf Xavier Bost}\\
With a background in Humanities, Xavier Bost began taking interest in computer science when attempting to automate processes used in historical linguistics. He is interested in language and image understanding working as a Ph.D. student at the \textsc{lia} (University of Avignon, France)  on automatic moving picture summarization.\\

\includegraphics[width=1in,clip,keepaspectratio]{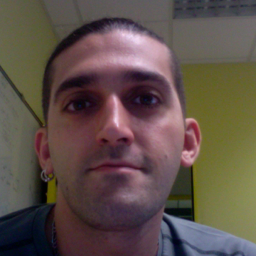}
\noindent {\bf Gr\'egory Senay}\\
Gr\'egory Senay received his Ph.D. in 2011 from \textsc{lia} at the University of Avignon, France and then worked as a postdoctoral fellow for two years. His research interests are in the areas of automatic speech recognition, spoken document retrieval, keyword spotting and topic classification with emphasis on speech confidence measure and audio indexing.\\

\includegraphics[width=1in,clip,keepaspectratio]{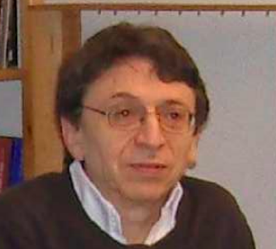}
\noindent {\bf Marc El-B{\`e}ze}\\
Marc El-B{\`e}ze is a full professor at the University of Avignon (\textsc{uapv}), France. He is an expert in several fields related to natural language processing: automatic text categorization, question answering, opinion analysis, and recommender systems. With some of his colleagues, he has participated to many evaluation campaigns such as \textsc{trec, clef, duc,} and more recently to RepLab. He has also been responsible for the French part of the European project EuroWordNet.\\

\includegraphics[width=1in,clip,keepaspectratio]{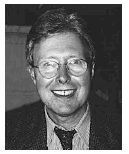}
\noindent {\bf Renato De Mori }\\
Renato De Mori is emeritus professor at Mc Gill University (Canada) and at the University of Avignon (France). He is a Fellow of the Computer Society and has been distinguished lecturer of the Signal Processing Society of the Institute of Electrical and Electronic Engineers (\textsc{ieee}). He is actually an Associate Editor of the \textsc{ieee} Transactions on Audio, Speech and Language.

His major contributions have been in the area of Automatic Speech Recognition and Understanding, Signal Processing, Computer Arithmetic, Software Engineering and Human-Machine Interfaces.\\
\end{document}